%% file: 2022_md_slam.tex
\documentclass[conference]{ieeeconf}
\IEEEoverridecommandlockouts      
\usepackage[table, x11names]{xcolor}
\usepackage[colorlinks=true,linkcolor=black,citecolor=black,urlcolor=blue]{hyperref}
\usepackage{times} 
\usepackage{amsmath} 
\usepackage{amssymb}  
\usepackage{graphicx}
\usepackage{algorithm}
\usepackage{mathtools}
\usepackage{algorithm}
\usepackage{algorithmicx}
\usepackage{float}
\usepackage{booktabs}
\usepackage[noend]{algpseudocode}
\usepackage[font=footnotesize]{caption}
\usepackage[font=footnotesize]{subcaption}
\usepackage[outdir=./]{epstopdf}
\usepackage{flushend}

\definecolor{brewerGreen0}{HTML}{E5F5F9}
\definecolor{brewerGreen1}{HTML}{99D8C9}
\definecolor{brewerGreen2}{HTML}{2CA25F}

\definecolor{brewerCyan0}{HTML}{ECE2F0}
\definecolor{brewerCyan1}{HTML}{A6BDDB}
\definecolor{brewerCyan2}{HTML}{1C9099}

\definecolor{brewerGrey0}{HTML}{F0F0F0}
\definecolor{brewerGrey1}{HTML}{BDBDBD}
\definecolor{brewerGrey2}{HTML}{636363}

\definecolor{revisionColor}{HTML}{0238A8} 
\definecolor{lastRevisionColor}{HTML}{CC4C02} 

\usepackage{cite}

\usepackage{balance}
\usepackage{multirow} 

\usepackage{glossaries-extra}
\setabbreviationstyle[acronym]{long-short}
\glssetcategoryattribute{acronym}{nohyperfirst}{true}

\makeglossaries

\newacronym{slam}{SLAM}{Simultaneous Localization and Mapping}
\newacronym{sfm}{SfM}{Structure from Motion}
\newacronym{pgo}{PGO}{Pose-Graph Optimization}
\newacronym{vpr}{VPR}{Visual Place Recognition}
\newacronym{sgd}{SGD}{Stochastic Gradient Descent}
\newacronym{ils}{ILS}{Iterative Least-Squares}
\newacronym{icp}{ICP}{Iterative Corresponding Point}
\newacronym{gn}{GN}{Gauss-Newton}
\newacronym{lm}{LM}{Levenberg-Marquardt}
\newacronym{pcg}{PCG}{Preconditioned Conjugate Gradient}
\newacronym{map}{MAP}{Maximum-A-Posteriori}
\newacronym{gf}{GF}{Gaussian Filters}
\newacronym{pf}{PF}{Particle Filters}
\newacronym{sdp}{SDP}{Semi-Definite Programming}
\newacronym{bst}{BST}{Binary Search Tree}
\newacronym{ndt}{NDT}{Normal Distributed Transform}
\newacronym{ba}{BA}{Bundle Adjustement}

\def\secref#1{Sec.~\ref{#1}}
\def\figref#1{Fig.~\ref{#1}}
\def\tabref#1{Tab.~\ref{#1}}
\def\eqref#1{Eq.~(\ref{#1})}

\def\ie{{i.e.}}

\def\lidar{LiDAR}
\def\lidars{LiDARs}
\def\rgbd{RGB-D}

\input{notation}

\newcounter{todonum}

\usepackage{lipsum}

\title{\LARGE \bf MD-SLAM: Multi-cue Direct SLAM}

\author{Luca Di Giammarino \and Leonardo Brizi \and Tiziano Guadagnino
  \and Cyrill Stachniss \and Giorgio Grisetti
  \thanks{Luca Di Giammarino, Leonardo Brizi, Tiziano Guadagnino and Giorgio Grisetti are
    with the Department of Computer, Control, and Management
    Engineering "Antonio Ruberti", Sapienza University of Rome, Italy,
    Email:\,\,{\tt\footnotesize{\{digiammarino, brizi, guadagnino,
        grisetti\}@diag.uniroma1.it.}}}%
   \thanks{Cyrill Stachniss is with the University of Bonn,
   	Germany, and also with the Department of Engineering Science at the University of Oxford, UK. Email:\,\,{\tt\footnotesize{cyrill.stachniss@igg.uni-bonn.de.}}
  }%
\thanks{We thank Gianmarco Roggiolani and Ignacio Vizzo for their precious support.}
}

\begin{document}
	\maketitle
	\thispagestyle{empty}
	\pagestyle{empty}
		

\begin{abstract}  
  \gls{slam}
  systems are fundamental building blocks for any autonomous
  robot navigating in unknown environments. The SLAM implementation heavily depends on the sensor modality employed on the
  mobile platform. For this reason, assumptions on the scene's structure are often made to maximize estimation accuracy.
  This paper presents a novel direct 3D SLAM pipeline that works
  independently for  \rgbd~and \lidar~sensors.
  Building upon prior work on multi-cue photometric frame-to-frame alignment \cite{della2018general}, our proposed approach provides an
  easy-to-extend and generic SLAM system. Our pipeline requires only minor adaptations within the projection model to handle different sensor modalities. We couple a position tracking system with
  an appearance-based relocalization mechanism that handles
  large loop closures. Loop closures are validated by the same direct registration algorithm used for odometry estimation. We present
  comparative experiments with state-of-the-art approaches on
  publicly available benchmarks using \rgbd~cameras and 3D
  \lidars. Our system performs well in heterogeneous datasets compared to other sensor-specific methods while making no assumptions about the environment. Finally, we release an open-source C++ implementation of our
  system.
  
\end{abstract}
	
\section{Introduction}
\label{sec:intro}
\gls{slam} is a popular field in robotics, and after roughly three
decades of research, effective solutions are available. As many sectors rely on SLAM, such as autonomous driving, augmented reality and space exploration, it still receives much attention in academia and industry. The advent of robust machine learning systems
allowed the community to enhance purely geometric maps with semantic
information or replace hard-coded heuristics with data-driven
ones. Within the computer vision community, we have seen photometric
(or direct) approaches used to tackle the \gls{slam} (or
Structure from Motion) problem. The direct techniques address
pairwise registration by minimizing the pixel-wise error between
image pairs. By not relying on specific features and having the
potential of operating at subpixel resolution on the entire image,
direct approaches do not require explicit data association and offer
the possibility to boost registration accuracy \cite{schops2019bad}. Whereas these methods
have been successfully used on monocular, stereo, or \rgbd~ images,
their use on 3D \lidar{} data is less prominent---probably due to
the comparably limited vertical resolution relation to cameras. Della Corte \emph{et al.} \cite{della2018general}
presented a multi-cue photometric registration methodology for \rgbd~cameras. It is a system that extends photometric approaches to different projective models and enhances the robustness by considering additional cues such as normals and depth or range in the cost
function. Recently released 3D \lidars{} sensors offer up to
128 beams, making direct approaches also more attractive for \lidar~data. In addition, most \lidars{} provide intensity or
reflectivity information besides range data. This intensity can be
used to sense a light reflectivity cue from the objects in the
environment. Being able to assemble an intensity-like image out of a
\lidar{} scan has unleashed the possibility of using well-known
computer vision appearance-based methods for place
recognition~\cite{di2021visual}.

\begin{figure}[t]
  \centering
  \includegraphics[width=1\linewidth]{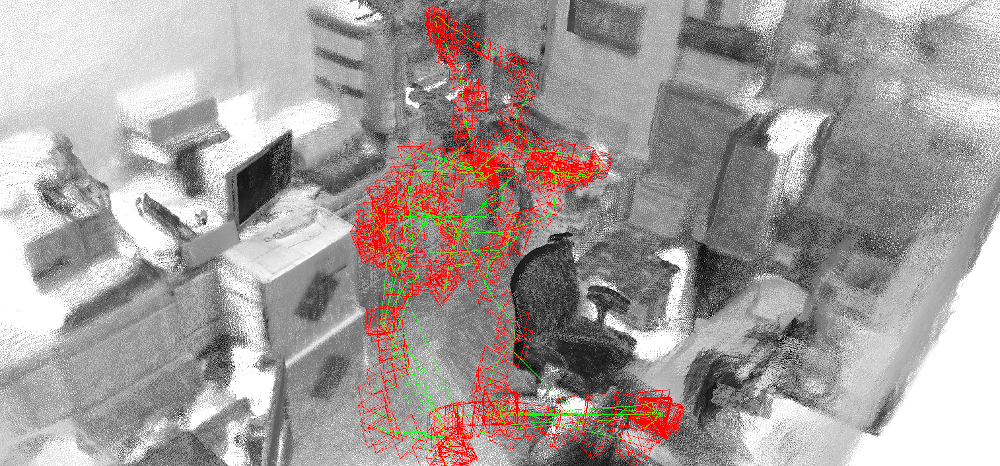}
  \vspace{0.1cm}\\
  \includegraphics[width=0.99\linewidth]{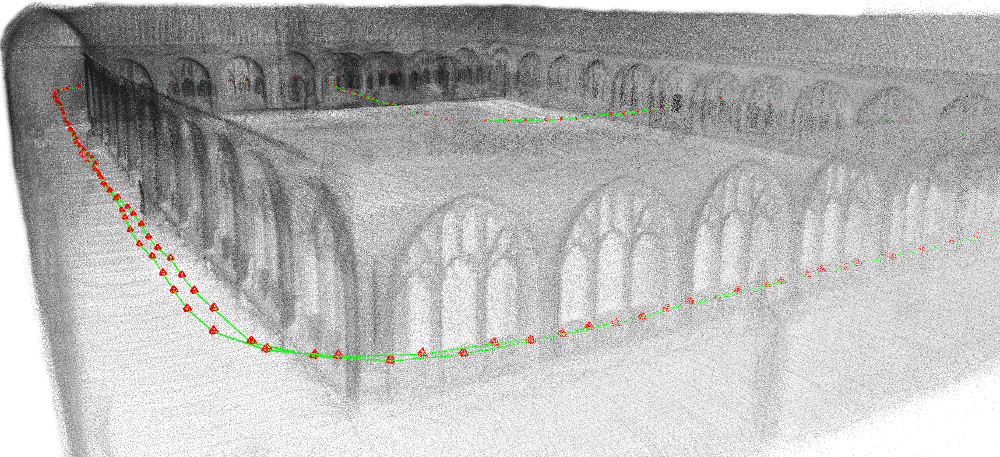}
  \caption{Scenes reconstructed using our pipeline. Top: results of a self-recorded dataset using Intel Realsense 455 \rgbd. Bottom: using \lidar~OS0-128 of the \textit{cloister} sequence from the Newer College Dataset \cite{zhang2021multicamera}.}
  \label{fig:lidar-data}
\end{figure}

The dominant paradigm for modern SLAM systems today is 
graph-based SLAM. A graph-based SLAM system works by constructing a
SLAM graph where each node represents the sensor position or a
landmark, while edges encode a relative displacement between
nodes. Pose-graphs are a particular case in which only poses are stored
in the graph. These local transformations stored in the edges are
commonly inferred by comparing and matching sensor readings. This
paper investigates the fusion of multi-cue direct registration
with graph-based SLAM.

The main contribution of this paper is a flexible, direct SLAM
pipeline for 3D data. To the best of our
knowledge, our approach is the only open-source SLAM
system that can deal with \rgbd~and \lidar~in a unified
manner. We realized a revised version of MPR \cite{della2018general} for computing the incremental motion of the sensor operating on \rgbd~as well as \lidar~data. We detect loop closures by an
appearance-based algorithm that uses a \gls{bst} structure proposed by Schlegel \emph{et al.}~\cite{schlegel2018hbst}, populated with binary feature descriptors \cite{rublee2011orb}. All components that require the
solution of an optimization problem rely on the same
framework~\cite{grisetti2020least}, resulting in a compact implementation. It is designed for flexibility, hence not optimizing the SLAM system to a specific sensor. Our system has been
tested on both, \rgbd{} and \lidar{} data, using benchmark
datasets. The accuracy is competitive concerning other
sensor-specific SLAM systems, while it outperforms them if some
assumptions about the structure of the environment are violated. An
open-source C++ implementation complements this work \footnote{https://github.com/digiamm/md\_slam}. \figref{fig:lidar-data}
illustrates example maps map built with our system using \rgbd~(top) and \lidar~(bottom).

\section{Related Work}
\label{sec:related}
3D \gls{slam} has been widely addressed by the computer vision and
robotics community and a large number of valid SLAM systems are
available. Whereas many deserve mention, we can focus only on some seminal works due to limited space in this section.
The available computational resources limited early approaches to operate
offline \cite{nuchter2005heuristic} in
fairly limited environments\cite{williams2007real}. After the
Kinect sensor became available about 15 years ago, we observed a
revamped interest in \rgbd~SLAM. Newcombe et
al. \cite{newcombe2011kinectfusion} were the first to leverage a dense
tracking on a Truncated Signed Distance Function (TSDF) stored in the GPU while using massively parallel implementation to render the
surface of the local scene perceived by the sensor. Meanwhile,
Segal~\emph{et al} \cite{segal2009generalized} proposed a robust
variant of \gls{icp} relying on a point-to-plane metric.  These
initial methods addressed the open-loop registration approaches,
tracking the pose of the sensor in a small neighbourhood. The advent
of efficient optimization systems such as iSAM~\cite{kaess2008isam}
and g2o~\cite{grisetti2011g2o}, made it possible to build an effective
full-fledged 3D SLAM system supporting loop closures and providing an
online globally consistent estimate. Novel efficient salient
floating point~\cite{bay2006surf} and binary image
descriptors~\cite{rublee2011orb}, paired with bag-of-words retrieval
methods inspired from web search engines, lead to impressive place
recognition approaches~\cite{galvez2012bags}. These methods were then
employed within visual \gls{slam} systems, ORB-SLAM by Mur
Artal~\emph{et. al}~\cite{mur2017orb} being one of the most
popular ones. The pipeline fully relied on the stability of features
(keypoints), minimizing the reprojection error of the reconstructed landmarks within the image. In contrast to these
indirect methods, another line of research aimed at photometric error
minimization. Keller \emph{et al.} \cite{keller2013pointfusion} use
projective data matching in a dense model, relying on a surfel-based
map for tracking. Others rely on keyframe-based technique
\cite{kerl2013dense}. As it happened for feature-based approaches,
these works were assembled into full visual SLAM
systems~\cite{engel2014lsd}. More recently, BAD-SLAM, a surfel-based
direct \gls{ba} system that combines photometric and geometric error
\cite{schops2019bad} using feature-based loop closures, shows that,
for well-calibrated data, dense \gls{ba} outperforms sparse \gls{ba}.
The accuracy and elegance shown by photometric approaches lead to
further developments such as MPR~\cite{della2018general} aiming at
unifying both \lidar~and \rgbd~devices into a unique registration method.

In parallel, the community approached \lidar-based odometry by seeking
alternative representations for the dense 3D point clouds.  These
include 3D salient features \cite{zhang2014loam,serafin2016fast},
subsampled clouds \cite{velas2016collar} or
\gls{ndt}~\cite{stoyanov2012fast}.  Nowadays, \lidar~Odometry and
Mapping (LOAM) is perhaps one of the most popular methods for \lidar~odometry \cite{zhang2014loam, zhang2015visual}. It extracts distinct
features corresponding to surfaces and corners, then used to
determine point-to-plane and point-to-line distances to a voxel
grid-based map representation. A ground optimized version (Lego-LOAM) method has been later proposed~\cite{shan2018lego}, as it leverages the presence of a ground plane in its segmentation and optimization steps. In contrast to sparse methods, dense approaches suffer less in a non-structured environment \cite{behley2018efficient}.
Compared to \rgbd~images, 3D \lidars~offer lower support for
appearance-based place recognition. It is common for dense
\lidar~SLAM systems to attempt a brute force registration with all neighbourhood clouds to seek loop closures. Thanks to the typically small drift, this strategy is most successful; however, computational costs grow significantly in large environments.
\lidar~loop closures have been addressed in different ways
compared to \rgbd. Magnusson \emph{et al.} proposed an approach
suitable for \gls{ndt}
representations~\cite{magnusson2009automatic}. R\"{o}hling et
al. \cite{rohling2015fast} investigated the use of histograms computed
directly from the 3D point cloud to define a measure of the similarity
of two scans. Novel types of descriptors have been investigated,
exploiting additional data gathered by the \lidar~sensor – \ie, light
emission of the beams \cite{cop2018delight, guo2019local}. However,
despite being very attractive, these descriptors are time-consuming to
extract and match, resulting in a slower system overall. Recent works
address loop-closures detection in a \rgbd~fashion, relying on the
visual feature matching extracted from the image obtained by using the
\lidar~intensity channel \cite{di2021visual}.

Building on top of prior work \cite{della2018general, di2021visual}, this paper presents a flexible and general SLAM approach. It is a direct method working on \rgbd and 3D \lidar data alike providing a unified approach. Our results show that it is competitive with other sensor-specific systems.
        

	
\begin{figure}[t]
	\centering
	\includegraphics[width=0.97\columnwidth]{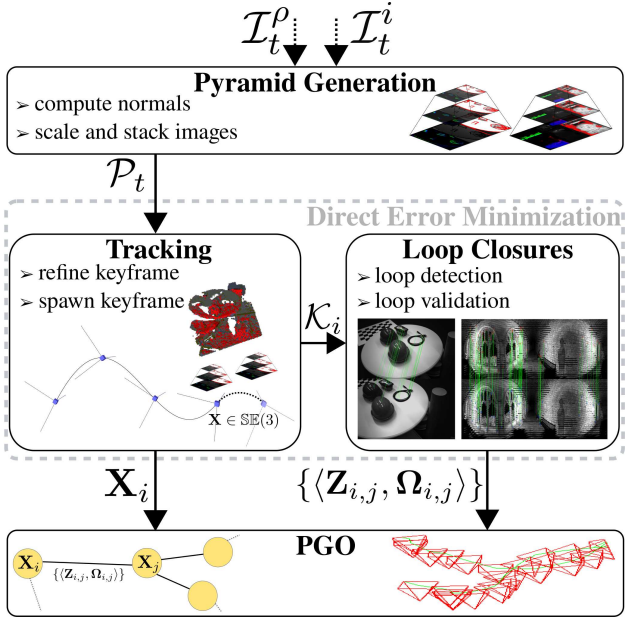}
	\caption{Illustration of our system. Range $\I_t^\mathrm{\rho}$ and $\I_t^\mathrm{i}$ images are taken as input from the system. An optimized trajectory within a map is produced as output. This system works independently both for \rgbd~and \lidar.}
	\label{fig:parameval}
\end{figure}
	
\section{Basics}
In this section, we outline some basic concepts used in
multiple modules of our system. The incremental position tracking (\secref{sec:tracking}), loop closure validation
(\secref{sec:closures}), and pose-graph solution (\secref{sec:pgo})
build upon \gls{ils}. All these modules are built on top
of the same software framework~\cite{grisetti2020least}. Our system
generalizes on range sensors by supporting different projective
models. In the remainder, we shortly describe how an \gls{ils} solution can be
found and recall projective models for \rgbd~and \lidars.
        
\subsection{Iterative Optimization}
\label{sec:iterative-optimization}
A generic Least Squares problem is captured by the following equation
\begin{equation}
	\label{eq:gen_error}
		\bx^* = \argmin_\bx \sum_k \| \be_k(\bx_k)
                \|_{\bOmega_k}^2.
	\end{equation}
Here $\be_k(\bx_k)$ is the error of the $k^\mathrm{th}$ measurement,
which is only influenced by a subset $\bx_k \in \bx$ of the overall
state vector $\bx$ and $\| \cdot \|^2_{\bOmega}$ represents the
squared Mahalonobis distance. \gls{ils} solves the above
problem by refining a current solution $\bx^*$.  At each iteration they construct a local quadratic approximation of
\eqref{eq:gen_error}:
\begin{equation}
          \sum_k \| \be_k(\bx^*_k + \bDeltax_k) \|_{\bOmega_k}^2
          \simeq \bDeltax^T \bH \bDeltax + 2 \bb^T \bDeltax + c.
        \end{equation}
The quadratic form is obtained by locally linearizing the vector error
term $\be_k(\bx_k)$ around the current solution, and assembling the
coefficients as follows:
\begin{align}
          &\be_k(\bx^*_k + \bDeltax_k) \simeq \underbrace{\be_k(\bx_k^*)}_{\be_k} +  \underbrace{ \frac{\partial \be_k(\bx_k)}{\partial \bx_k} }_{\bJ_k}  \bDeltax_k,
          \label{eq:gen-linearization}\\
          &\bb=\sum_k {\bJ_k^T} \bOmega_k \be_k, \qquad \bH = \sum_k {\bJ_k^T} \bOmega_k \bJ_k 	  \label{eq:Hb-linear} .
\end{align}
The minimum of the quadratic form is then found as the solution
$\bDeltax^*$ of the linear system $\bH\bDeltax^* = \bb$. The computed
perturbation is finally applied to the current solution $\bx^*
\leftarrow \bx^* + \bDeltax^*$. This procedure is iterated until
convergence.

Should the state be a smooth manifold $\bX \neq \bbR^n$, the problem
admits a local Euclidean parameterization $\bDeltax$ on a chart
constructed around $\bX^*$.  In this case, the Taylor expansion of
\eqref{eq:gen-linearization} is evaluated at the origin
$\bDeltax=\bZero$ of the chart computed around the current estimate
$\bX^*$.  Once a new perturbation vector $\bDeltax^*$ is obtained by
solving the linear system, the estimate is updated through the boxplus
operator $\bX^* \leftarrow \bX^* \boxplus \bDeltax^*$ as reported in
\cite{grisetti2020least}.

All modules in our system carry on optimization on one or more
variables in $\bbSE(3)$, represented as homogeneous transformation
matrices. As perturbation for the optimization, we use $\bDeltax \in \bbR^6$. This encodes translation and the
imaginary part of the normalized quaternion.  We define the $\boxplus$
operator $\bX' = \bX \boxplus \bDeltax = \bX \cdot \vTot(\bDeltax)$,
as a function that applies the tranform obtained from perturbation $\v2t(\bDeltax)$ to the transform $\bX$. Similarly, we define
the operator boxminus $\boxminus$ as the one that calculates the vector
perturbation between two manifold points as $\bDeltax =
\bX'\boxminus\bX = \tTov(\bX' \bX^{-1})$.

\subsection{Projections}
A projection is a mapping $\pi : \bbR^3 \rightarrow \Gamma \subset
\bbR^2$ from a world point $\bp = [x, y, z]^T$ to image coordinates
$\bu = [u, v]^T$.  Knowing the depth or the range $\rho$ of an image
point $\bu$, we can calculate the inverse mapping $\pi^{-1} : \Gamma
\times \bbR \rightarrow \bbR^3$, more explicitly $\bp = \pi^{-1}(\bu,
\rho)$. We will refer to this operation as unprojection.  In the
remainder, we recall the \emph{pinhole} projection that models with
\rgbd~cameras, and the spherical projection that captures 3D \lidars.

\textbf{Pinhole Model:} Let $\bK$ be the camera matrix. Then, the
pinhole projection of a point $\bp$ is computed as
\begin{eqnarray}
	\pi_p(\bp) &=& \phi(\bK \, \bp)\\
	\bK&=&\begin{bmatrix}
	f_x& 0 & c_x\\
	0 & f_y & c_y\\
	0 & 0 &1
	\end{bmatrix} \label{eq:camera-matrix-p}\\
	\phi(\bv) &=& \frac{1}{v_z}
	\begin{bmatrix}
	v_x \\ v_y
	\end{bmatrix}
	\label{eq:pinhole-projection},
\end{eqnarray}
with the intrinsic camera parameters for the focal length~$f_x$, $f_y$
and the principle point~$c_x$, $c_y$. The function~$\phi(\bv)$ is the
homogeneous normalization with $\bv=[v_x,v_y,v_z]^T$.
	
\textbf {Spherical Model:} Let $\bK$ be a camera matrix in the form
of~\eqref{eq:camera-matrix-p}, where $f_x$ and $f_y$ specify
respectively the resolution of azimuth and elevation and $c_x$ and
$c_y$ their offset in pixels. The function $\psi$ maps a 3D point to
azimuth and elevation. Thus the spherical projection of a point is
given by
\begin{eqnarray}
	\pi_s(\bp) &=& \bK_{[1,2]} \psi(\bp) \\
	\psi(\bv) &=& 
	\begin{bmatrix}
	\atantwo(v_y, v_x) \\
	\atantwo\left(v_z, \sqrt{v_x^2 + v_y^2}\right)\\ 1
	\end{bmatrix},
	\label{eq:spherical-projection}
\end{eqnarray}
Note that in the spherical model $\bK_{[1,2]} \in \bbR^{2 \times 3}$,
being the third row in $\bK$ suppressed.
        
\section{Our Approach}
\label{sec:main}
Our approach relies on a pose-graph to represent the map. Nodes of the
pose-graph store keyframes in the form of multi-cue image
pyramids. Our pipeline takes as input intensity and depth images for \rgbd~or intensity and range images for \lidar. For compactness, we will generalize, mentioning only range images. The pyramids are generated from the inputs
images each time a new frame becomes available. By processing the range information, our system computes the surface normals
and organizes them into a three-channel image, which is then stacked to the
original input to form a five-channel image. Pyramids are generated by
downscaling this input. This process is described in
\secref{sec:pyramids}.

The pyramids are fed to the tracker, which is responsible for
estimating the relative transform between the last keyframe and the
current pyramid through the direct error minimization strategy
summarized in \secref{sec:error-min}. The tracker is in charge of
spawning new keyframes and adding them to the graph when necessary, as
discussed in \secref{sec:tracking}.

Whenever a new keyframe is generated, the loop closure schema,
described in \secref{sec:closures}, seeks for potential relocalization
candidates between the past keyframes by performing a search in
appearance space. Candidate matches are further pruned by geometric
validation and direct refinement. Successful loop closures result in
the addition of new constraints in the pose-graph and trigger a complete
graph optimization as detailed in \secref{sec:pgo}.
    
\begin{figure}
  \includegraphics[width=\columnwidth]{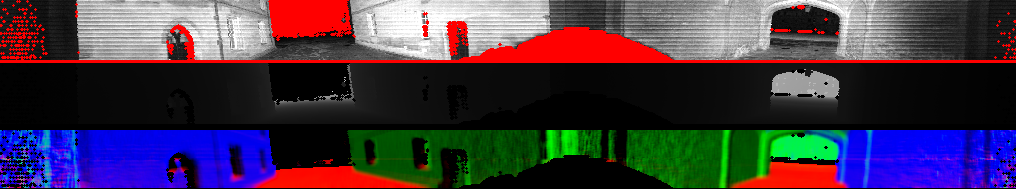}
  \vspace{0.1cm}\\
  \includegraphics[width=\columnwidth]{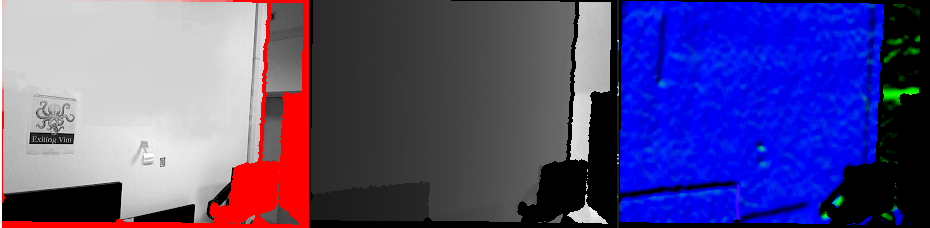}

  \caption{Cues generated for \lidar~(top) and \rgbd~(bottom) images.  The first row/column shows the intensity $\I^i$, the middle shows the range $\I^\mathrm{\rho}$, and the last one illustrates the normals encoded by color $\I^\mathrm{n}$. The red pixels on the intensity cues are invalid measurements (i.e., range not available).
    }
  \label{fig:cues}
\end{figure}
 
\subsection{Pyramid Generation}
\label{sec:pyramids} 
The first step to generate a pyramid from a pair of intensity
$\I^\mathrm{i}$ and range image $\I^\mathrm{\rho}$ consists of extracting the normals. To calculate the normal at pixel $\bu$ we
unproject the pixels in the neighborhood $\mathcal U=\{\bu':
\|\bu-\bu'\| < \tau_\bu\}$ whose radius $\tau_\bu$ is inversely proportional to the
range at the pixel $I^{\rho}(\bu)$. The normal $\bn_\bu$ is the one of
the plane that best fits the unprojected points from the set $\mathcal
U$ . All valid normals are assembled in a normal image
$\I^\mathrm{n}$, so that $\I^\mathrm{n}(\bu)=\bn_\bu$.

One level of a pyramid $\I$, therefore, consists of three images:
$\I^\mathrm{i}$, $\I^\mathrm{\rho}$ and $\I^\mathrm{n}$. Further
channels such as curvature and semantics can easily be embedded into the representation by adding additional images. In the remainder, we will
refer to one general image in the set $\I$ as a \emph{cue}
$\I^\mathrm{c}$.

Pyramids are required to extend the basin of convergence of
direct registration methods. This is due to the implicit data
association, which operates only in the neighborhood of a few pixels.
Hence, downscaling the images increases the convergence basin at the cost of reduced accuracy. However, the accuracy can be recovered by running the registration from the coarser to the finest level. Each time a level is changed, the initial guess of
the transformation is set to the solution of the previous coarser
level.

A pyramid $\cP$ is generated from all the cues $\I=\{\I^\mathrm{c}
\}$, by downscaling at user-selected resolutions. In our experiments,
we typically use three scaling levels, each of them half the
resolution of the previous level.
    
\subsection{Direct Error Minimization}
\label{sec:error-min}
As in direct error minimization approaches, our method seeks to find
the transform $\bX^*\in \bbSE(3)$ that minimizes the photometric
distance between the two images:
\begin{equation}
      \begin{small}
        \bX^*=
	\argmin_{\bX \in \bbSE(3)} \sum_{\bu}
	\| 
	\underbrace{
	  \predI^\mathrm{i}(\bu) - \I^\mathrm{i}\overbrace{\left(\pi \left(\bX \pi^{-1} \left(\bu,\rho \right) \right) \right)}^{\bu'}}_{\be_{\bu}} \|^2
      \end{small}
      \label{eq:total-error}
\end{equation} 
Where $\be^\mathrm{i}_\bu$ denotes the error between corresponding
pixels. The evaluation point $\bu'$ of $\I^\mathrm{i}$ is computed by
unprojecting the pixel $\bu$, applying the transform $\bX$ and
projecting it back. To carry out this operation, the range at the pixel
$\rho = \I^{\rho}(\bu)$ needs to be known.

\eqref{eq:total-error} models classical photometric error
minimization assuming that the cues are not affected by the transform $\bX$.
In our case, range and normal are affected by $\bX$. Hence, we need
to account for the change in these cues, and we will do it by
introducing a mapping function $\map^\channel(\bX,
\predI^\channel(\bu))$. This function calculates the \emph{pixel}
value of the $\channel^\mathrm{th}$ cue after applying the transform
$\bX$ to the original channel value $\predI^\channel(\bu)$.  We can
thus rewrite a more general form of \eqref{eq:total-error} that
accounts for all cues and captures this effect as follows:
\begin{equation}
	 \bX^*= \argmin_{\bX \in \bbSE(3)} \sum_ \channel \sum_\bu
	\| 
	\underbrace{
	  \map^\channel(\bX,\predI^\channel(\bu)) - \I^\channel(\bu')}_{\be_{\bu}^c}\|_{\bOmega^\channel}^2
        \label{eq:total-error-multicue}
\end{equation}
The squared Mahalanobis distance $\| \cdot \|^2_{\bOmega^c}$ is used
to weight the different cues. More details about a general
methodology for direct registration can be found in
\cite{della2018general}.

While approaching the problem in \eqref{eq:total-error-multicue} with the \gls{ils} method described in
\secref{sec:iterative-optimization}, particular care has to be taken
to the numerical approximations of floating-point numbers.  In
particular, since each pixel and cue contribute to
constructing the quadratic form with an independent error
$\be_{\bu}^c$, the summations in \eqref{eq:Hb-linear} might accumulate
millions of terms.  Hence, to lessen the effect of these round-offs,
\eqref{eq:Hb-linear} has to be computed using a stable algorithm. In
our single-threaded implementation, we use the compensated summation algoritm~\cite{higham2002accuracy}.

We use multi-cue direct alignment in incremental position tracking,
explained in next section (\secref{sec:tracking}) and in loop closure
refinement and validation (\secref{sec:closures}).

\subsection{Tracking}
\label{sec:tracking}
This module is in charge of estimating the open-loop trajectory of the
sensor. To this extent, it processes new pyramids as they become
available by determining the relative transform between the last
pyramid $\cP_t$, and the current keyframe $\cK_i$. A keyframe stores a
global transform $\bX_i$, and a pyramid $\cP_i$. The registration
algorithm of \secref{sec:error-min} is used to compute a relative
transform $\bZ_{i,t}$ between the last two pyramids. Whenever the
magnitude of such a transform exceeds a given threshold or the overlap
between $\cP_i$ and $\cP_t$ becomes too small, the tracker spawns a
new keyframe $\cK_{i+1}$, with transform $\bX_{i+1}=\bX_{i+1}
\bZ_{i,i+1}$. Furthermore, it adds to the graph a new constraint
between the nodes $i$ and $i+1$, with transform $\bZ_{i,i+1}$, and
information matrix $\bOmega_{i,i+1}$. The latter is set to $\bH$
matrix of the direct registration at the optimum. The generation of
the new keyframe triggers the loop detection described in the next
section. Using keyframes reduces the drift that would occur when
performing subsequent pairwise registration since the reference frame
stays fixed for a longer time. Potentially, if the sensor hovers at a
distance smaller than the keyframe threshold, all registrations are
done against the same pyramid, and no drift would occur.

\subsection{Loop Detection and Validation}
\label{sec:closures}
This module is responsible for relocalizing a newly generated keyframe
with respect to previous ones. More formally, given a query frame
$\cK_i$, it retrieves a set of tuples $\{\left < \cK_j, \bZ_{i,j},
\bOmega_{i,j}\right>\}$, consisting of a past keyframe $\cK_j$, a
transform $\bZ_{i,j}$ between $\cK_i$ and $\cK_j$ and an
information matrix $\bOmega_{i,j}$ characterizing the uncertainty of
the computed transform.  Our system approaches loop closing in multiple
stages. At first, we carry on visual place recognition on the
intensity channels. This approach leverages the results of previous
work \cite{di2021visual}. For visual place recognition, we rely on ORB
feature descriptors, extracted from the $\I^\mathrm{i}$ of each
keyframe. Retrieving the most similar frame to the current one
results in looking for the images in the database having the closest descriptor ``close'' to the one of the current image. To
efficiently conduct this search, we use a hamming distance embedding
binary search tree (HBST)~\cite{schlegel2018hbst}, a tree-like
structure that allows for descriptor search and insertion in
logarithmic time by exploiting particular properties of binary
descriptors. A match from HBST also returns a set of pairs of
corresponding points between the matching keypoints. Having the depth
and unprojecting the points, we can carry on a straightforward RANSAC
registration. Finally, each candidate match is subject to direct
refinement (\secref{sec:error-min}). This step enhances the
accuracy and it provides information matrices on
the same scale as the ones generated by the tracker. The above strategy is applied independently to \rgbd~or \lidar~data.
These surviving pairs $\{\left < \cK_j, \bZ_{i,j},
\bOmega_{i,j}\right> \}$, constitute potential loop closing
constraints to be added to the graph.  However, to handle environments
with large sensor aliasing, we introduced a further check to 
preserve topological consistency. Whenever a loop closure is found, we
carry on a direct registration between all neighbours that would
result \emph{after} accepting the closure. If the resulting error is within certain bounds, the
closure is finally added to the graph, and a global optimization is
triggered.

\subsection{Pose-graph Optimization}
\label{sec:pgo}
The goal of this module is to retrieve a configuration of the
keyframes in the space that is maximally consistent with the
incremental constraints introduced by the tracker and the loop closing
constraints by the loop detector.  A pose-graph is a special case of a
factor graph~\cite{grisetti2020least, kaess2008isam}.  The nodes of
the graph are the keyframe poses $\bX=\{ \bX_i\}_{i=1:N}$, while the
constraints encode the relative transformations between the connected
keyframes, together with their uncertainty $\{ \left< \bZ_{i,j},
\bOmega_{i,j} \right> \}$.  Optimizing a factor graph consists in
solving the following optimization problem:
\begin{equation}
\bX^*=
\argmin_{\bX \in \bbSE(3)^N} \sum_{i,j} \| \underbrace{ \bX_i^{-1} \bX_j \boxminus \bZ_{i,j} }_{\be_{i,j}}  \|^2_{\bOmega_{i,j}}
\label{eq:pgo-error}
\end{equation} 
Here, the error $\be_{i,j}$ is the difference between predicted
displacement $\bX_i^{-1} \bX_j$ and result of the direct alignment
$\bZ_{i,j}$.  The total perturbation vector
$\bDeltax \in \bbR^{6N}$ results from stacking all variable perturbations $\{\bDeltax_i\}$.

\begin{figure}[t]
	\centering
	\centering
	\includegraphics[width=0.40\textwidth]{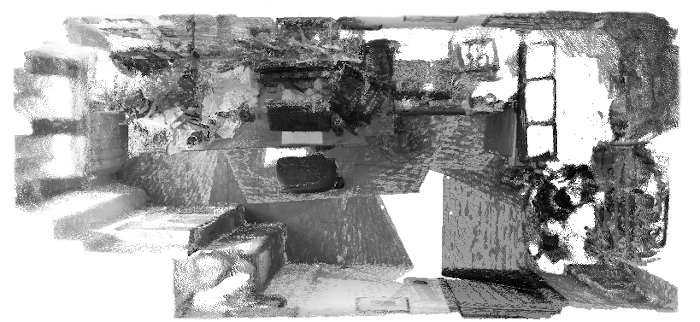}
	\centering
	\includegraphics[width=0.40\textwidth]{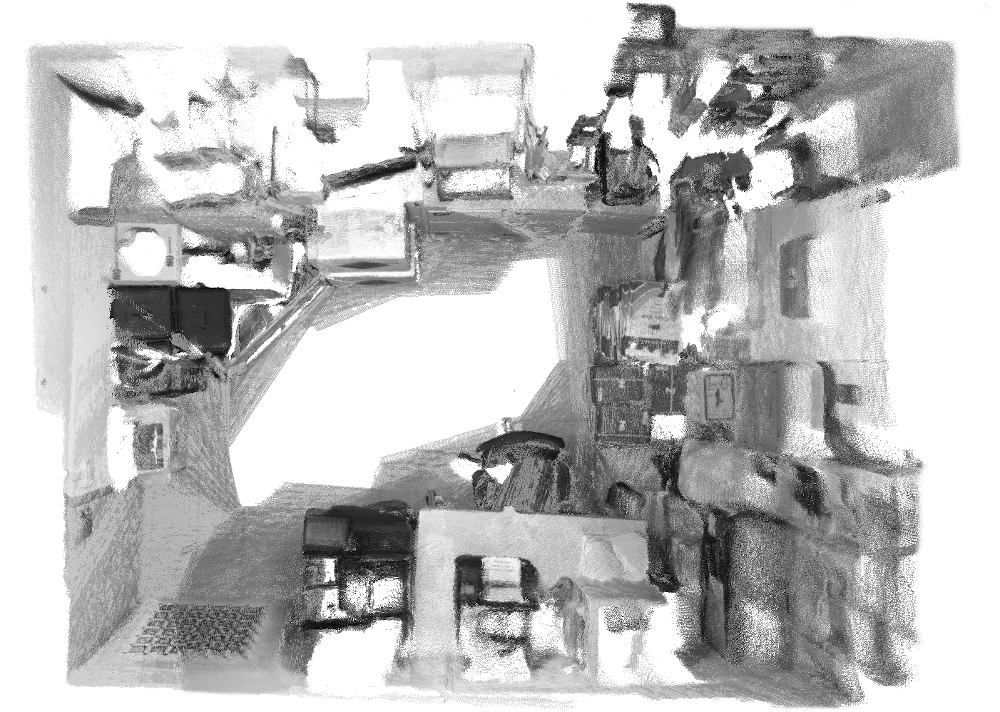}
	\caption{Qualitative \rgbd~reconstructions showing the global consistency produced by our pipeline. Data has been self-recorded with an Intel Realsense 455.}
	\label{fig:rgbd}
\end{figure}

\section{Experimental Evaluation}
\label{sec:exp}
In this section, we report the results of our pipeline on different
public benchmark datasets. To the best of our knowledge, our approach
is the only open-source SLAM system that can deal with
\rgbd~and~\lidar~in a unified manner. Therefore, to evaluate our
system, we compare with state-of-the-art SLAM packages developed specifically
for each of these sensor types. For \rgbd~we consider DVO-SLAM
\cite{kerl2013dense} and ElasticFusion \cite{whelan2015elasticfusion}
as direct approaches and ORB-SLAM2 \cite{mur2017orb} as indirect
representative. For \lidar~we compare against LeGO-LOAM
\cite{shan2018lego} as feature-based and SuMA
\cite{behley2018efficient} representing the dense category.

To run the experiments, we used a PC with an Intel Core i7-7700K CPU @
4.20GHz and 16GB of RAM. Since this work is focused on SLAM, we perform our quantitative evaluation using the RMSE on the absolute trajectory error (ATE) with $\bbSE(3)$ alignment. The alignment for the metric is computed by using the Horn method \cite{horn1988closed}, and the
timestamps are used to determine the associations. Then, we calculate the RMSE of
the translational differences between all matched poses.

The tracking module dominates the runtime of our approach since
loop closures are detected and validated asynchronously within
another thread. Hence, we report the average
frequency at which the tracker runs for each sensor. At the core of the tracker, we
have the photometric registration algorithm, whose computation is
proportional to the size of the images. Despite our current
implementation of the registration algorithm being single-threaded, on
the PC used to run the experiments, the tracking system runs at 5 Hz for the sensor with the highest resolution, while it can operate online on the sensor with the lowest resolution.

\begin{figure}
	\centering
	\centering
	\includegraphics[width=0.95\linewidth]{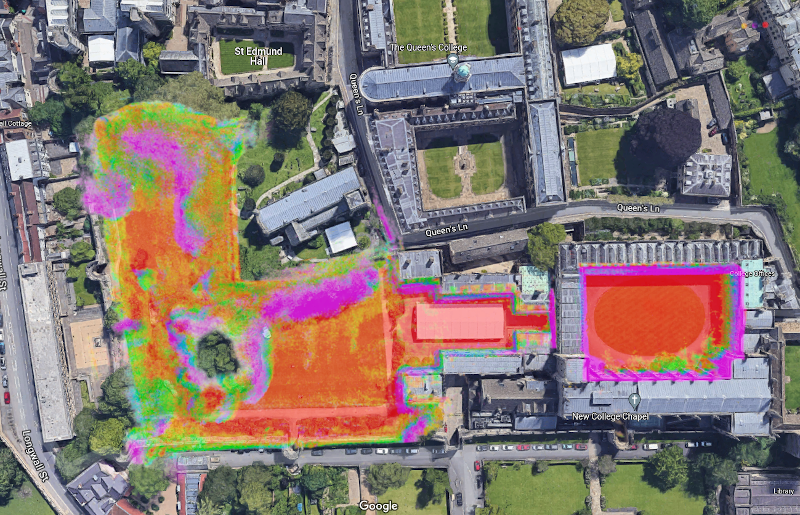}
	\caption{MD-SLAM map on \textit{long} sequence from Newer College Dataset \cite{ramezani2020newer} aligned with Google Earth.}
	\label{fig:gmaps}
\end{figure}

\subsection{\rgbd~Results}
We conducted several experiments with \rgbd~sensor. Qualitative analysis have been done using self-recorded data and are shown in \figref{fig:rgbd}.
As public benchmarks we used the TUM-\rgbd~\cite{sturm2012benchmark} and
the ETH3D~\cite{schops2019bad}.
The TUM \rgbd~dataset contains multiple real datasets
captured with handheld Xbox Kinect. A rolling
shutter camera provides RGB data. Further, the camera’s depth and color streams are not
synchronized. Every sequence accompanies an accurate groundtruth
trajectory obtained with an external motion capture system.
ETH3D benchmark is acquired with global shutter cameras
and accurate active stereo depth. Color and depth images are
synchronized. We select several indoor sequences for which
ground-truth, computed by external motion capture, is available.

On these datasets, we compare with DVO-SLAM, ElasticFusion and
ORB-SLAM2. These three approaches are representative of different
classes of SLAM algorithms.  \tabref{tab:rgbd-fr} shows the results on
the TUM \rgbd~datasets, while \tabref{tab:rgbd-eth} presents
the outcome on the ETH3D datasets.  DVO SLAM implements a mixed
geometry-based and direct registration. Internally the alignment
between pairs of keyframes is obtained by jointly minimizing
point-to-plane and photometric residuals. This is similar to
ElasticFusion, whose estimate consists of a mesh model of the
environment and the current sensor location instead of the
trajectory. In contrast to these two approaches, ORB-SLAM2 implements
a traditional visual SLAM pipeline, where a local map of landmarks
around the \rgbd~sensor is constructed from ORB features. This map is
constantly optimized as the camera moves by performing local \gls{ba}. Loop closures are detected through DBoW2
\cite{galvez2012bags} and a global optimization on a
$\mathbb{S}\mathrm{im}(3)$ pose-graph to enforce global consistency is
used.

The TUM dataset provides images $640\times 480$ pixels, while ETH3D $740\times
460$ pixels. From these images, we compute a 3 level pyramid with
scales $1/2$, $1/4$, and $1/8$. Our system runs respectively at $5.5$ and $5$ Hz at these resolutions.

\newcolumntype{L}{>{$}l<{$}}
\newcolumntype{C}{>{$}c<{$}}
\newcolumntype{R}{>{$}r<{$}}
\newcommand{\nm}[1]{\textnormal{#1}}

\begin{table} [t]
	\centering
	\begin{tabular}{LCCC}
		\toprule
		&
		\multicolumn{1}{c}{fr1/desk} &
		\multicolumn{1}{c}{fr1/desk2}     &
		\multicolumn{1}{c}{fr2/desk} \\
		\midrule
		\nm{DVO-SLAM} & 0.021 & 0.046 & 0.017  \\
		\nm{ElasticFusion} & 0.020 & 0.048 & 0.071  \\
		\nm{ORB-SLAM2} & \textbf{0.016} & \textbf{0.022} & \textbf{0.009}  \\
		\nm{\textbf{Ours}} & 0.041 & 0.064 & 0.057  \\
		\bottomrule
	\end{tabular}
	\caption{ATE RMSE [m] results on
		TUM RGB-D datasets, recorded with non-synchronous
		depth using a rolling shutter camera.}
	\label{tab:rgbd-fr}
\end{table}

\begin{table} [t]
	\centering
	\begin{tabular}{LCCCCCC}
		\toprule
		&
		\multicolumn{1}{c}{table3} &
		\multicolumn{1}{c}{table4}     &
		\multicolumn{1}{c}{table7} &
		\multicolumn{1}{c}{cables1} &
		\multicolumn{1}{c}{plant2} &
		\multicolumn{1}{c}{planar2} \\
		\midrule 
		\nm{DVO-SLAM} & 0.008 & 0.018 & \textbf{0.007} & \textbf{0.004} & 0.002 & 0.002 \\
		\nm{ElasticFusion} & - & 0.012 & - & 0.018 & 0.017 & 0.011\\
		\nm{ORB-SLAM2} & \textbf{0.007} & \textbf{0.008} & 0.010 & 0.007 & 0.003 & 0.005\\
		\nm{\textbf{Ours}} & 0.021 & 0.022 & 0.036 & 0.015 & \textbf{0.001} & \textbf{0.001}\\
		\bottomrule
	\end{tabular}
	\caption{ATE RMSE [m] on
		ETH3D, recorded with  global shutter camera and synchronous streams. ElasticFusion fails in \textit{table3} and \textit{table7}.}
	\label{tab:rgbd-eth}
\end{table}

In \tabref{tab:rgbd-fr} we can see that ORB-SLAM2 clearly outperforms
all other pipelines. DVO-SLAM and ElasticFusion provide 
comparable results, and our approach is the worst in terms of
accuracy. Yet, the largest error is 6.4 cm, which results in a usable
map. As stated before, this dataset is subject to rolling shutter and
asynchronous depth effects. ORB-SLAM2, being feature-based, is less
sensitive to these phenomena. DVO-SLAM and ElasticFusion explicitly
model these effects. Our approach does not attempt to address these
issues since it would render the whole pipeline less consistent
between different sensing modalities. 

\tabref{tab:rgbd-eth} presents the results on the ETH3D benchmark. 
In this case, our performances are on par with other methods, since
intensity and depth are synchronous, and the camera is global shutter.

These results highlight the strength and weaknesses of a purely
direct approach not supported by any geometric association. While being compact, it suffers from unmodeled effects and requires a considerable overlap between subsequent frames.

\subsection{3D \lidar~Results}
We conducted different experiments on public \lidar~ benchmarks to show the performances of our SLAM implementation. 
For the \lidar~we use the Newer College
Dataset~\cite{ramezani2020newer, zhang2021multicamera} recorded at
10 Hz with two models of Ouster \lidars: OS1 and OS0.  We conducted our
evaluation on the \emph{long}, \emph{cloister}, \emph{quad-easy} and
\emph{stairs} sequences. The OS1 has 64 vertical beams.  We selected the \emph{long} sequence that lasts
approximately 45 minutes.  It consists of multiple loops with
viewpoint changes between buildings and a park.

The other three shorter sequences are recorded with the OS0, which has
128 vertical beams.  The \lidar~ \emph{quad-easy}
sequence contains four loops that explore quad, \emph{cloister} mixes outdoor and indoor scenes while \emph{stairs} is purely indoor
and based on vertical motion through different floors.

\begin{table} [h!]
	\centering
	\begin{tabular}{LCRCR}
		\toprule
		\multicolumn{1}{l}{} &
		\multicolumn{3}{c}{OS0-128} &
		\multicolumn{1}{c}{OS1-64}  \\ 
		\cmidrule{2-5}
		&
		\multicolumn{1}{c}{cloister} &
		\multicolumn{1}{c}{quad}     &
		\multicolumn{1}{c}{stairs} &
		\multicolumn{1}{c}{long}     \\
		\midrule
		\nm{LeGO-LOAM} & \textbf{0.20} & \textbf{0.09} & 3.20 & \multicolumn{1}{c}{1.30} \\
		\nm{SuMA} & 3.34 & 1.74 & 0.67 & \multicolumn{1}{c}{-} \\
		\nm{\textbf{Ours}} & 0.36 & 0.25 & \textbf{0.34} & \multicolumn{1}{c}{1.74} \\
		\bottomrule
	\end{tabular}
	\caption{ATE RMSE [m] results of all benchmarked approaches on the Newer College Dataset. SuMA fails on \textit{long} sequence.}
	\label{tab:lidar}
\end{table}

\begin{figure}
\centering
\begin{subfigure}{0.20\textwidth}
	\centering
	\includegraphics[width=\textwidth]{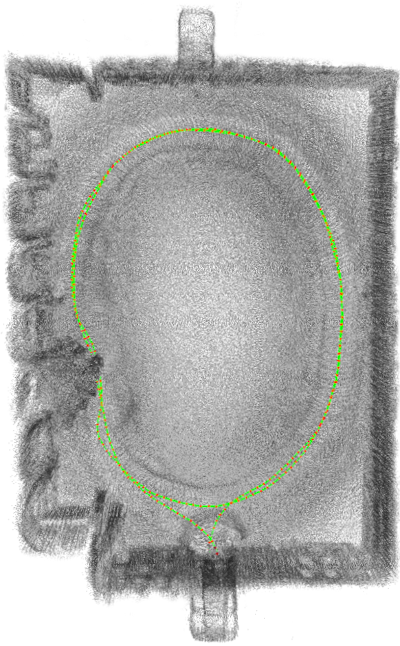}
	\caption{quad-easy}
	\label{fig:quad-easy}
\end{subfigure}
\hfill
\begin{subfigure}{0.27\textwidth}
	\centering
	\includegraphics[width=\textwidth]{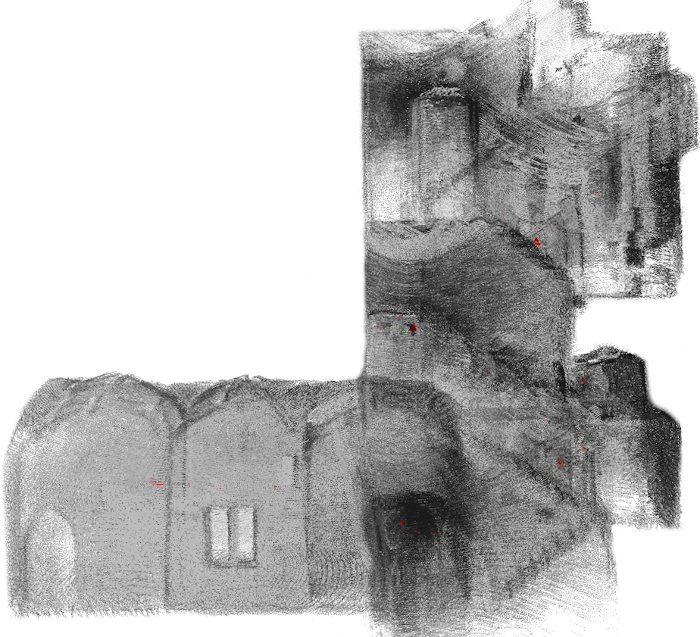}
	\caption{stairs}
	\label{fig:stairs}
\end{subfigure}
\caption{Some scenes from Newer College dataset reconstructed by our system.}
\label{fig:lidar}
\end{figure}

\begin{figure}[t]
	\centering
	\centering
	\includegraphics[width=0.95\linewidth]{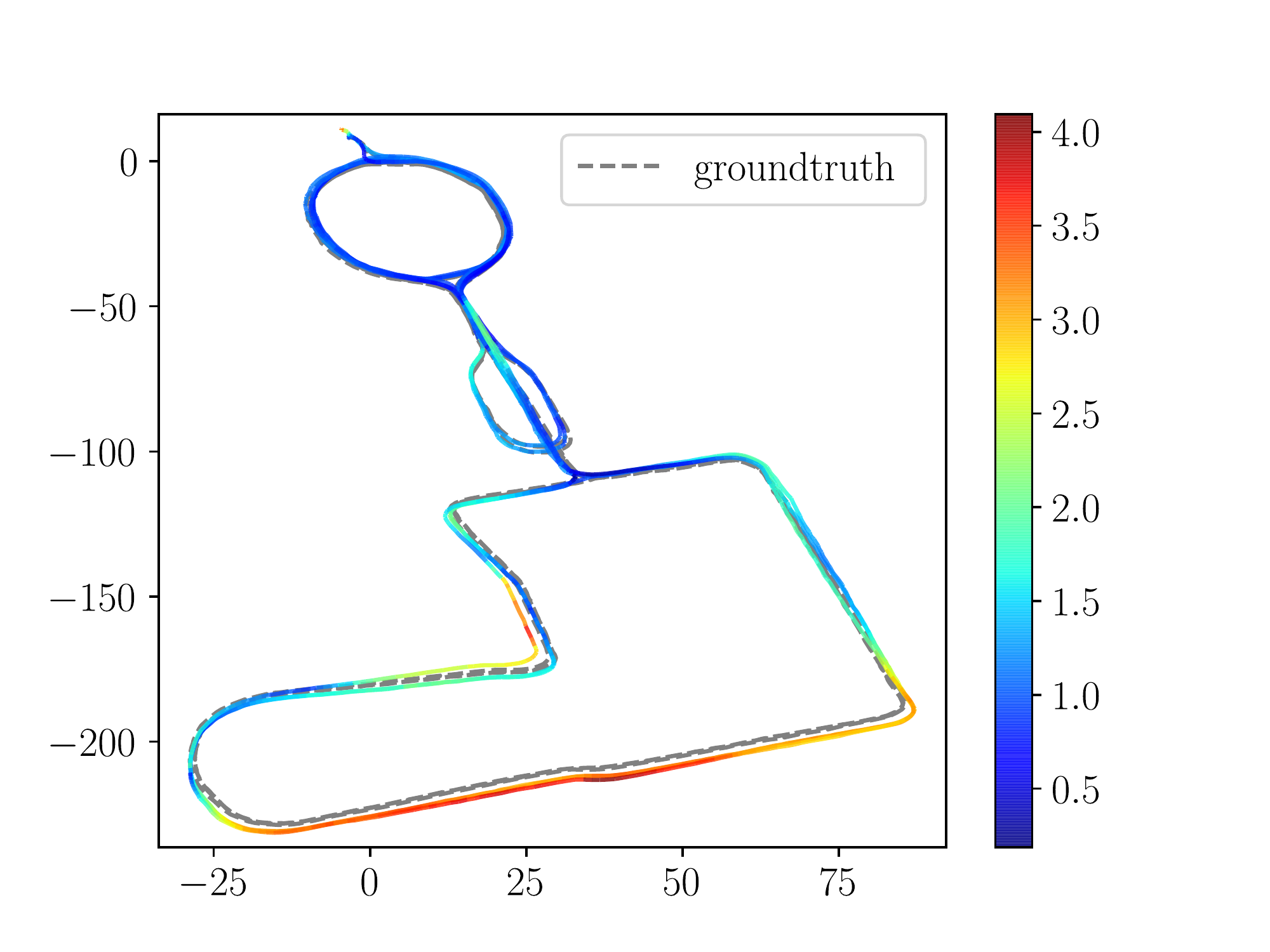}
	\caption{Alignment of our estimate with the groundtruth in \textit{long} sequence of Newer College. The color bar on the right shows the translational error [m]
		over the whole trajectory.}
	\label{fig:traj}
\end{figure}

Qualitative analysis have been performed to show the results obtained by our pipeline. \figref{fig:lidar} illustrates some reconstructions obtained with MD-SLAM from Newer College sequences. \figref{fig:gmaps} and \figref{fig:traj} show the global consistency of our estimate on \textit{long} sequence.

Quantitatively, we compare against LeGO-LOAM and SuMA. These
represent two different classes of \lidar~algorithms, respectively
sparse and dense. \tabref{tab:lidar} summarizes the results of the
comparison. LeGO-LOAM is currently one of the most accurate
\lidar~SLAM pipelines and represents a sparse class of
\lidar~algorithms. In contrast to our approach, LeGO-LOAM is a pure
geometric feature-based frame-to-model \lidar~SLAM work, where the
optimization on roll, yaw and z-axis (pointing up) is decoupled from
the planar parameters. SuMa constructs a surfel-based map and estimates
the changes in the sensor’s pose by exploiting the projective data
association in a frame-to-model or in a frame-to-frame
fashion. For both the pipelines loop closures are handled through ICP. Being ground optimized, LeGO-LOAM shows impressive results
mainly in chunks where ground occupies most of the scene, yet our
approach provides competitive accuracy. The situation becomes
challenging for LeGO-LOAM when its assumptions are violated, such as in
the \emph{stairs} sequence. In this case, our pipeline is the most
accurate since it does not impose any particular structure on the
environment being mapped. SuMA performances are the worst in terms of
accuracy. We tried this pipeline both in a frame-to-frame and
frame-to-model mode. The one reported in \tabref{tab:lidar} represents SuMA frame-to-frame that always outperforms the frame-to-model on these datasets.

We use the OS1 to produce images of $64 \times 1024$ pixels while the OS0 to produce images of $128 \times 1024$ pixels.
Since the horizontal resolution is much larger than the vertical one,
to balance the aspect ratio for direct registration, initially, we
downscale the horizontal resolution by $1/2$ for OS0 and by $1/4$ for
OS1. Our approach generates a pyramid with the following scales: $1$,
$1/2$ and $1/4$. With these settings, our system operates at around 10
Hz on the OS0 and at approximately 20 Hz on the OS1, making it suitable for
online estimation.

\section{Conclusion}
\label{sec:conclusion}
In this paper, we presented a direct SLAM system that operates both
with \rgbd~and \lidar~sensors. These two heterogeneous sensor
modalities are addressed exclusively by changing the projection models. To the best of our
knowledge, our approach is the only open-source SLAM
system that can deal with \rgbd~and \lidar~in a unified
manner. 
All optimization
components in our system are dealt with a single \gls{ils} solver, resulting in highly compact code.
Comparative experiments show that our generic method can compete with sensor-specific state-of-the-art approaches. Being purely photometric and without making any assumption of the environment, our pipeline shows consistent results on different types of datasets. 
We release our software as a C++ open-source package. The current
single-thread implementation can operate online with small image
sizes. Thanks to the inherent data-separation of direct
registration, we envision a GPU implementation of our approach that
seamlessly scales to high resolution while matching real-time
requirements. Furthermore, the independence of the internal representation
from the sensor source paves the way to SLAM systems that operate jointly
on both \rgbd~and \lidar.
\\

\bibliographystyle{plain}
\bibliography{2022_md_slam}

\balance

\end{document}

%% file: notation.tex
\usepackage{amsopn}

\newcommand{\bv}{\mathbf{v}}

\newcommand{\bK}{\mathbf{K}}

\newcommand{\bH}{\mathbf{H}}

\newcommand{\bX}{\mathbf{X}}
\newcommand{\bZ}{\mathbf{Z}}

\newcommand{\bJ}{\mathbf{J}}
\newcommand{\bZero}{\mathbf{0}}

\newcommand{\map}{\zeta}

\newcommand{\cK}{\mathcal{K}}
\newcommand{\cP}{\mathcal{P}}

\newcommand{\bbR}{\mathbb{R}}
\newcommand{\bbSE}{\mathbb{SE}}

\newcommand{\bb}{\mathbf{b}}

\newcommand{\be}{\mathbf{e}}

\newcommand{\bx}{\mathbf{x}}

\newcommand{\bu}{\mathbf{u}}
\newcommand{\bn}{\mathbf{n}}

\newcommand{\bp}{\mathbf{p}}

\newcommand{\bDeltax}{\mathbf{\Delta x}}

\newcommand{\tTov}{\mathrm{t2v}}
\newcommand{\vTot}{\mathrm{v2t}}

\newcommand{\bOmega}{\mathbf{\Omega}}

\DeclareMathOperator*{\argmin}{argmin}
\DeclareMathOperator*{\atantwo}{atan2}

\def\g2o{$g^2o$}
\def\t2v{\mathrm{t2v}}
\def\v2t{\mathrm{v2t}}
\def\ev2t{\mathrm{ev2t}}

\newcommand{\I}{\ensuremath{\mathcal{I}}}
\newcommand{\predI}{\ensuremath{\mathcal{\hat I}}}
\newcommand{\channel}{\mathrm{c}}